\documentclass[10pt,conference]{IEEEtran}
\IEEEoverridecommandlockouts

\usepackage{amsmath,amsfonts,bm}

\def\eqref#1{equation~\ref{#1}}

\def\1{\bm{1}}

\DeclareMathAlphabet{\mathsfit}{\encodingdefault}{\sfdefault}{m}{sl}
\SetMathAlphabet{\mathsfit}{bold}{\encodingdefault}{\sfdefault}{bx}{n}

\usepackage{cite}
\usepackage{amsmath,amssymb,amsfonts}
\usepackage{algorithmic}
\usepackage{graphicx}
\usepackage{textcomp}

\usepackage{hyperref}
\usepackage{url}
\usepackage{caption}
\usepackage{subcaption}
\usepackage[table]{xcolor}
\usepackage{booktabs}

\def\BibTeX{{\rm B\kern-.05em{\sc i\kern-.025em b}\kern-.08em
    T\kern-.1667em\lower.7ex\hbox{E}\kern-.125emX}}

\usepackage{tikz}
\newcommand\copyrighttext{%
  \footnotesize \textcopyright\ 2026 IEEE. Personal use of this material is permitted. Permission from IEEE must be obtained for all other uses, in any current or future media, including reprinting/republishing this material for advertising or promotional purposes, creating new collective works, for resale or redistribution to servers or lists, or reuse of any copyrighted component of this work in other works. 
  Published in: J. K. Suwa, J. P. Silveira e Silva and B. I. Grisci, ``Post-Hoc Feature Selection Layer for Neural Networks Interpretability,'' \textit{2026 International Joint Conference on Neural Networks (IJCNN)}, Maastricht, Netherlands, 2026, pp. 1--6, doi: \href{https://doi.org/10.1109/IJCNN69854.2026.11729834}{10.1109/IJCNN69854.2026.11729834}.}
\newcommand\copyrightnotice{%
\begin{tikzpicture}[remember picture,overlay]
\node[anchor=south,yshift=8pt] at (current page.south) {\fbox{\parbox{\dimexpr\textwidth-\fboxsep-\fboxrule\relax}{\copyrighttext}}};
\end{tikzpicture}%
}

\begin{document}

\title{Post-Hoc Feature Selection Layer for Neural Networks Interpretability
\thanks{This research  was funded by Fundaç\~ao de Amparo à Pesquisa do Estado do Rio Grande do Sul (\texttt{FAPERGS}) [24/2551-0000725-3; 25/2551-0002116-2].
This work was supported by the Serrapilheira Institute (\texttt{grant number Serra - R-2501-51351}).
This work was supported by PIBIC CNPq-UFRGS.
This study was financed in part by the Coordination for the Improvement of Higher Education Personnel – Brazil (CAPES) – Finance Code 001.}
}

\author{\IEEEauthorblockN{João Kenji Suwa}
\IEEEauthorblockA{\textit{Institute of Informatics} \\
\textit{UFRGS}\\
Porto Alegre, Brazil \\
joao.suwa10@gmail.com}
\and
\IEEEauthorblockN{João Pedro Silveira e Silva}
\IEEEauthorblockA{\textit{Institute of Informatics} \\
\textit{UFRGS}\\
Porto Alegre, Brazil \\
jpedross1999@gmail.com}
\and
\IEEEauthorblockN{Bruno Iochins Grisci}
\IEEEauthorblockA{\textit{Institute of Informatics} \\
\textit{UFRGS}\\
Porto Alegre, Brazil \\
bigrisci@inf.ufrgs.br}
}

\maketitle
\copyrightnotice

\begin{abstract}
The interpretability of complex neural networks remains a critical challenge, especially for models already deployed in high-stakes domains. 
To address this, we adapted the Feature Selection Layer (FSL) to optimize feature weights post-training, a variant we term post-hoc FSL.
Our approach reframes the FSL as a lightweight, trainable module that integrates with already frozen pre-trained models on tabular datasets to highlight the features the original model considers most important. 
This post-hoc FSL learns feature relevance by fine-tuning its weights based on the original model's learned outputs. 
Crucially, this process is non-invasive, operating without altering the original model's architecture or its learned parameters. 
We conducted our experiments using both statistical and visual metrics, including accuracy, F1 score, recall, precision, weighted t-SNE and silhouette score, and also analyzed the stability of the post-hoc FSL on high-dimensional synthetic and real-world tabular datasets. 
We compared the post-hoc FSL feature weighting method using these metrics against the original embedded FSL and other post-hoc interpretability methods, such as Integrated Gradients, Noise Tunnel, DeepLIFT, Gradient SHAP, and Feature Ablation. 
Experimental results demonstrate that the post-hoc FSL feature weighting method successfully identified relevant features across the different datasets, maintaining the predictive power of the original neural network while enhancing its interpretability. 
\end{abstract}


\section{Introduction}

Over the past decade, the availability of large-scale data has established Deep Neural Networks (DNNs) as state-of-the-art tools for knowledge extraction in domains such as computer vision and healthcare \cite{badawy2023healthcare,KHAN2023100026}. However, due to their lack of interpretability, their “black box” nature poses serious challenges for transparency, ethical accountability, and bias mitigation \cite{Unraveling_the_Black_Box:2024,prince2023understanding}.
Addressing the trade-off between accuracy and interpretability remains a critical challenge in the safe and reliable deployment of deep learning systems.
To address this interpretability problem, feature weighting methods assign relevance scores to input features, providing insights into feature behavior in DNN models \cite{molnar2020interpretable}.
Among these techniques, the Feature Selection Layer (FSL), proposed by \cite{figueroa2021towards}, is an embedded feature attribution method designed to learn feature relevance based on the model’s internal dynamics during the training phase. 
However, its embedded nature requires joint training from scratch, limiting its use for interpreting pre-trained DNN models. 
To overcome this limitation, we adapted the FSL to optimize feature weights after the initial training phase, a variant we term post-hoc FSL.
Post-hoc FSL preserves the efficacy of the original method while enabling feature relevance analysis on pre-trained networks.
This approach enhances interpretability and potentially improves predictive performance.
Furthermore, by keeping the backbone network frozen, this method improves computational efficiency, as the optimization of FSL weights converges rapidly compared to training from scratch.

\section{Related Work}

\subsection{Feature Selection}


Feature selection aims to identify a compact subset of relevant features to improve model performance and reduce complexity \cite{BARBIERI2024123667}.
Most approaches rely on feature weighting to assign relevance scores and are commonly categorized by their interaction with the learning algorithm \cite{tahir2007simultaneous,miao2016survey}.
Filter methods are model-agnostic and computationally efficient but often ignore feature interactions. 
Wrapper and embedded methods exploit a specific predictive model, achieving strong performance at the cost of higher computational expense and limited generalization across architectures \cite{BARBIERI2024123667}.
Hybrid methods combine multiple strategies to balance efficiency and accuracy \cite{ang2015supervised}, while ensemble methods aggregate multiple selection runs to improve robustness and stability.

\subsection{Feature Selection Layer}

The Feature Selection Layer (FSL), introduced by \cite{figueroa2021towards}, is an embedded feature weighting method for neural networks. 
It consists of a simple layer between the neural network's input and its first hidden layer, where each neuron maintains a one-to-one correspondence with an input feature. 
Weights, initialized as 1/$n$, where $n$ is the total number of features, are jointly trained with the network and reflect feature relevance during learning.
A modified L1 regularization term is added to ensure the interaction between features, which is calculated as:
\begin{equation}
r(W^{\text{FSL}}) = \lambda \cdot \left| \sum_{t=1}^{n} W_t^{\text{FSL}} - 1 \right|
\end{equation}
where $W^{\text{FSL}}$ represents the weights from the FSL, $n$ is the number of features and $\lambda$ is a value between 0 and 1 that determines the strength of the regularization.
FSL then adds two hyperparameters to the network: the FSL activation function and the $\lambda$ term for regularization.

\subsection{Post-hoc Interpretability Methods}
\label{subsec:posthocmethods}


Post-hoc feature attribution methods are essential for interpreting ``black box'' models such as neural networks.
Gradient-based approaches, including Integrated Gradients \cite{sundararajan2017axiomatic}, DeepLIFT \cite{shrikumar2017learning}, and Gradient SHAP \cite{lundberg2017unified}, estimate feature relevance relative to a reference input. Integrated Gradients and Gradient SHAP compute attributions along paths from the baseline to the input, while DeepLIFT propagates relevance based on differences from reference activations.
Other methods include Feature Ablation, a perturbation-based technique, and Noise Tunnel, which improves the robustness of gradient-based attributions by smoothing \cite{kokhlikyan2020captumunifiedgenericmodel}.
In our experiments, we apply Noise Tunnel with Integrated Gradients and compare the proposed post-hoc FSL method against these approaches using the metrics described in Section \ref{subsec:evaluation_metrics}.

\section{Proposed Method}
\label{headings}

We propose an adaptation of FSL designed to enhance the interpretability of pre-trained neural networks, which we term post-hoc FSL.
Our method adapts the FSL \cite{figueroa2021towards}, reformulating it as a trainable module applicable to any compatible pre-trained neural network.
The core objective is to identify the most relevant features for a model's predictions without altering its original learned parameters.

\subsection{Architecture and Mechanism}

\begin{figure}[ht]
    \centering
    \includegraphics[width=0.75\linewidth]{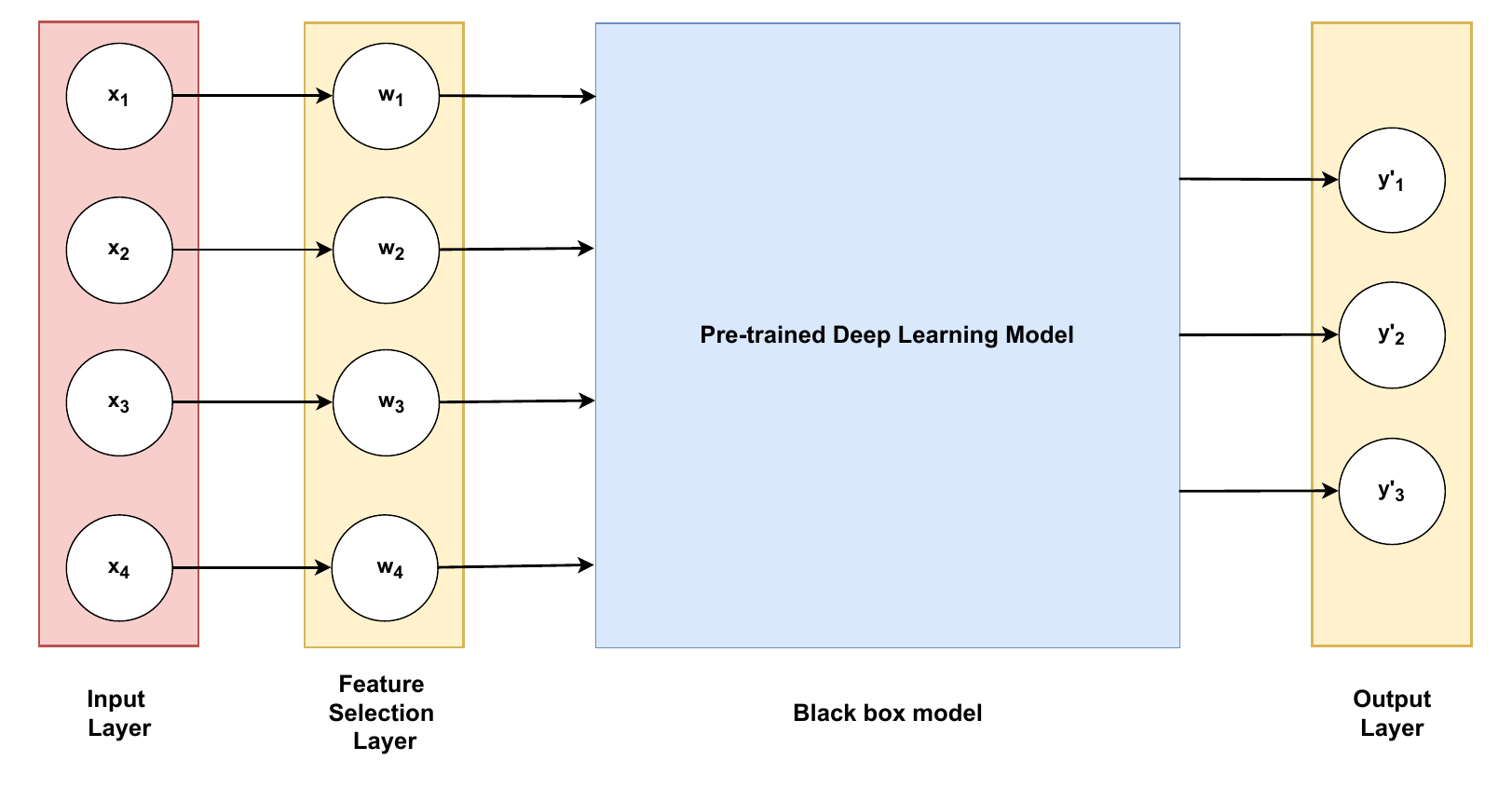}
    \caption[Representation of post-hoc Feature Selection Layer.
    Representation of post-hoc Feature Selection Layer.
    A neural network is shown with four inputs, followed by the FSL, a pre-trained ``black box'' deep learning model and three output neurons.
    Each input is connected to its weight node on FSL layer before being passed to the pre-trained model.
    The pre-trained model defined each of the tree outputs for this multiclass sample.
    During the backpropagation process, the FSL's weights are changed while the pre-trained model remains frozen.
    In the end of the process the FSL layer will learn the feature weights.]
    {\textbf{Representation of post-hoc Feature Selection Layer.} 
    In this example, four input features are individually multiplied by their activated weights within the FSL before being passed to the pre-trained model. 
    During fine-tuning, only the weights within the FSL are updated, while all parameters of the pre-trained model remain frozen.
    }
    \label{fig:posthoc_feature_selection_layer}
\end{figure}

Architecturally, the proposed post-hoc FSL remains faithful to the original FSL.
It is a simple, trainable layer with a one-to-one mapping between neurons and input features, placed between the input and a pre-trained network (\autoref{fig:posthoc_feature_selection_layer}). 
Each feature $\vec{x_i}$ is scaled by a trainable weight $\vec{w_i}$, producing a weighted vector $\vec{wx}$ passed to the subsequent layers of the network.
Unlike the original FSL, our method leverages the pre-learned internal weights of the network to highlight which features are considered most relevant, without modifying the model's internal parameters. 
After the analysis, the post-hoc FSL yields a set of feature weights that quantifies feature importance, while also being capable of improving predictive performance by amplifying relevant features and suppressing noise.
Importantly, the post-hoc FSL can be removed after the analysis, allowing the pre-trained model to revert to its original state.
Despite its distinct training paradigm, we retain the inherent benefits of the FSL architecture, which is computationally efficient and performs competitively in dimensionality reduction and accuracy when compared to AFS \cite{gui2019afs}, Random Forest, ReliefF, and Mutual Information \cite{figueroa2021towards}.

\subsection{Post-Hoc Training Procedure}

Our feature weighing approach operates via a distinct post-hoc training procedure. Given a pre-trained model $f_{\text{pre}}$ with its internal parameters frozen, a vector $\vec{x}$ that represents a single input, a vector $\vec{w}$ which represents the weights from the FSL, an activation function $a$, a loss function $\ell$, and a learning rate $\eta$, the training process is as follows:

\textbf{Step 1:} Our feature weighting layer is integrated between the input layer and the hidden layers of the pre-trained neural network (\autoref{eq:weights_and_inputs}).

\textbf{Step 2:} Data is fed through the combined model (FSL and the frozen network), and the final output is used to compute the loss function of the whole model (\autoref{eq:internal_model_input}).


\textbf{Step 3:} During backpropagation the parameters of the pre-trained network remain unchanged, while gradients are computed only for the FSL weights.


\textbf{Step 4:} Steps 2 and 3 are repeated over several epochs. This allows the weights of the post-hoc FSL to converge, with their final magnitudes representing the feature importance.
\begin{align}
  \vec{x}' &= a(\vec{w}^{(i)}) \odot \vec{x} \label{eq:weights_and_inputs} \\
  \vec{y}  &= f_{\text{pre}}(\vec{x}') \label{eq:internal_model_input}
\end{align}

\subsection{Layer Configuration: Regularization, Activation, and Initialization}

We adopt three design choices to regulate the Feature Selection Layer (FSL). 
First, we apply the standard $L1$ regularization to promote sparsity by shrinking weights of irrelevant features toward zero. 
Second, we use ReLU activation to ensure non-negative weights, simplifying interpretation: a weight of zero indicates irrelevance, and positive values represent features' importance. 
Third, we initialize weights to $1.0$ instead of $1/n$, providing a stable starting point in which all features are ingested by the pre-trained model initially using unchanged values and preventing initializations with weights close to zero. 

\section{Experiments}
\label{subsec:experiments}

This section details the experimental setup used to evaluate the proposed post-hoc FSL method. 
We describe the datasets (Section \ref{subsec:datasets}), the evaluation metrics (Section \ref{subsec:evaluation_metrics}), and the specific implementation details (Section \ref{subsec:experiments_setup}).

\subsection{Datasets}
\label{subsec:datasets}

To evaluate the proposed feature selection layer, we employed synthetic datasets with predefined relevant and noisy features (Section \ref{subsubsec:synthdatasets}) and real-world datasets, including email spam classification \cite{balaka2020spam} and high-dimensional microarray data with limited samples \cite{cumida:2019} (Section \ref{subsubsec:realdatasets}).
Using diverse datasets allows us to evaluate post-hoc FSL’s effectiveness in reducing dimensionality and improving neural network interpretability by identifying relevant features and suppressing noise.
All datasets consist of complete numerical features with no missing values and are fully described in \autoref{table:datasets}.

\begin{table}[ht]
\caption{\textbf{Datasets used for the experiments}}
\label{table:datasets}
\begin{center}
\resizebox{\columnwidth}{!}{
\begin{tabular}{llllllll}
\multicolumn{1}{c}{\bf Name}  &\multicolumn{1}{c}{\bf Feature type} &\multicolumn{1}{c}{\bf Labels} &\multicolumn{1}{c}{\bf Features} &\multicolumn{1}{c}{\bf Informative} &\multicolumn{1}{c}{\bf Folds} &\multicolumn{1}{c}{\bf Samples} &\multicolumn{1}{c}{\bf Test Split}
\\ \hline \\
XOR &Binary &2 &50 &2 &11 &500 &20\% \\
SynthA &Numerical &2 &100 &30 &11 &3000 &10\%\\
Spam &Numerical &2 &3000 &Unknown &11 &5172 &20\%\\
Liver-22405 &Numerical &2 &22284 &Unknown &15 &48 &10\%\\
Breast-45827 &Numerical &6 &54676 &Unknown &15 &151 &10\%\\
\end{tabular}
}
\end{center}
\end{table}

\subsubsection{Synthetic Datasets}
\label{subsubsec:synthdatasets}

For this analysis, we used the XOR and SynthA synthetic datasets, with controlled proportions of relevant and irrelevant features. 
SynthA was generated by first defining an $n$-dimensional hypercube, where $n$ is the number of informative features. 
Then, Gaussian clusters of data points are created at each vertex of the hypercube, with an equal number of clusters assigned to each class. 
The informative features are the ones that are coordinates for the created points in the hypercube \cite{BARBIERI2024123667}.
Synthetic data provides ground truth for feature relevance, enabling direct evaluation of post-hoc FSL's selection accuracy (Section \ref{subsec:evaluation_metrics}), based on the weights assigned to each feature.

\subsubsection{Real-world datasets}
\label{subsubsec:realdatasets}

To evaluate the post-hoc FSL on real-world high-dimensionality and low-sample-size challenges, we selected datasets from two domains. 
First, we utilized microarray datasets from the CuMiDa repository \cite{cumida:2019}, a database of cancer datasets selected from over $30,000$ GEO experiments, following bias-reduction guidelines by \cite{grisci2024use}. 
We used Liver-GSE22405
(binary classification) and Breast-GSE45827 
(six-class classification). 
Additionally, we evaluated a spam email dataset \cite{balaka2020spam}, represented by word frequencies in the email corpus.

\subsection{Evaluation Metrics}
\label{subsec:evaluation_metrics}


To evaluate the post-hoc FSL, we employed statistical, visual, and stability metrics (Sections \ref{subsubsec:statisticalmetrics}, \ref{subsubsec:visualmetrics}, and \ref{subsec:stability}) to assess its effectiveness.

\subsubsection{Performance Metrics}
\label{subsubsec:statisticalmetrics}

We assess model predictive performance using accuracy, precision, recall, and F1 score. 
Given a synthetic dataset with known ground truth and a subset of top-ranked features, we evaluate feature selection using two metrics \cite{BARBIERI2024123667} ranging from $0.0$ to $1.0$: the percentage of informative features selected, \textbf{PIFS} (\autoref{equationPIFS}), and the percentage of selected features that are informative, \textbf{PSFI} (\autoref{equationPSFI}).

\begin{align}
  \mathit{PIFS} = \frac{\lvert S_{\text{selected}} \cap S_{\text{informative}} \rvert}
                     {\lvert S_{\text{informative}} \rvert} \label{equationPIFS} \\
  \mathit{PSFI} = \frac{\lvert S_{\text{selected}} \cap S_{\text{informative}} \rvert}
                     {\lvert S_{\text{selected}} \rvert} \label{equationPSFI}
\end{align}

\subsubsection{Visual Metrics}
\label{subsubsec:visualmetrics}

For visual analysis, we use a variant of t-SNE (t-Distributed Stochastic Neighbor Embedding) \cite{maaten2008visualizing}, a technique that projects high-dimensional data into a low-dimensional space (typically two dimensions) based on pairwise distances between points.
To visualize the discriminative power of post-hoc FSL features, we use weighted t-SNE \cite{grisci2025assessing}. 
This algorithm scales Euclidean distances by feature importance, ensuring that highly relevant dimensions contribute more significantly to the final spatial arrangement.
Higher values indicate well-separated and dense clusters, while values near $0$ suggest boundary points and values close to $-1$ indicates that the points are closer to some other cluster than the one they are assigned.
The primary objective of using weighted t-SNE is to provide a global visual inspection tool to compare how different feature scoring techniques impact data clustering.


\subsubsection{Stability Metrics}
\label{subsec:stability}

To evaluate the stability of the post-hoc FSL feature weighting method, we adopted a stability analysis protocol based on data perturbation, which assesses how the method’s output varies under small changes in the training data \cite{awada2012review}.
We employed $k$-fold stratified cross-validation, where the dataset is partitioned into $k$ folds and the feature weighting method is applied $k$ times, each time using $k$–1 folds for training and one fold for validation.
This process produces $k$ sets of feature weights, which are then analyzed to assess stability by measuring their consistency across folds using three metrics proposed by \cite{BARBIERI2024123667}.
The Jaccard Index measures the overlap between two sets with the top-$n$ selected features, capturing the subset stability.
The Pearson correlation evaluates the similarity of feature weights.
Lastly, the Spearman rank correlation quantifies the agreement between feature rankings.

\subsection{Experiments setup}
\label{subsec:experiments_setup}

In our experimental evaluation, we trained 10 models, being two per dataset:
(1) a baseline model without embedded feature weighting and
(2) a model with an embedded Feature Selection Layer (FSL).
The baseline models were designed to be interpreted by the proposed post-hoc FSL and compared post-hoc methods (Section \ref{subsec:posthocmethods}). All models employed different architectures and were trained using the Adam optimizer (learning rate $0.001$) with a weighted cross-entropy loss.
To assess the effectiveness of the post-hoc FSL, we compared it with state-of-the-art post-hoc methods and the original FSL using the statistical, visual, and stability metrics described in Section \ref{subsec:evaluation_metrics}, computed as the average across cross-validation folds.
Additionally, we extended the evaluation to TabPFN, a pre-trained transformer-based neural network \cite{hollmann2022tabpfn}.


\section{Results and Discussion}

\subsection{Results for Synthetic Datasets}
\label{subsec:synthresults}

When comparing the predictive performance for the XOR dataset, the baseline model achieved near-optimal predictive performance, $0.999$ across all metrics, while both FSL and post-hoc FSL reached perfect scores of $1.0$. 
All models consistently identified the two relevant features across all folds, achieving perfect PIFS, PSFI, and Jaccard similarity scores of $1.0$ when evaluated on the top-$2$ features. This consistency is further supported by the Pearson correlation metric, which yielded values near $1.0$.
Despite this, the Spearman rank correlation remained near 0.0 across all methods, indicating that while relevant features were prioritized over irrelevant ones, their relative rankings can vary.
When comparing the silhouette scores and weighted t-SNE visualizations, all models achieved statistically similar results, while improving cluster separability when compared to the standard t-SNE ($p < 0.01$).

For the SynthA dataset, both FSL variants surpassed the baseline across all predictive metrics (F1 score, accuracy, precision and recall). Post-hoc FSL showed moderate improvement, yielding scores of approximately $0.916$, whereas FSL consistently delivered the highest performance, with scores approaching $0.954$.
Furthermore, both FSL variants and the alternative post-hoc methods yielded statistically comparable results in terms of cluster separability (with all methods significantly outperforming the baseline, $p < 0.01$).
Despite this statistical parity, numerical variations were observed in the average silhouette scores.
Post-hoc FSL recorded the lowest average silhouette score ($0.115$), contrasting with higher averages for FSL ($0.191$) and the other post-hoc approaches, which ranged from a minimum of $0.181$ (Feature Ablation) to a maximum of $0.235$ (Integrated Gradients).
Regardless of these constraints, the FSL post-hoc reached PFSI and PIFS scores of $0.906$ on the top-$30$ features, demonstrating its ability to identify informative features, albeit in this specific case, less effectively than the FSL, which reached perfect scores of $1.0$ on both metrics.
In comparison, the other post-hoc approaches achieved scores in the range of $0.96$–$0.99$.
These results influenced post-hoc FSL stability scores
(\autoref{table:synthstabilityresults}), which also had an inferior performance when compared to the other methods.
We hypothesize that the synthetic dataset's structure may have hindered post-hoc FSL's feature identification. 
Nonetheless, its performance on real-world, high-dimensional datasets remained strong.


\begin{table}[ht]
\caption{
\textbf{Results for stability metrics using the SynthA dataset. }
Bold indicates the highest average.
Jaccard was calculated using the top-$30$ weighted features.
}
\label{table:synthstabilityresults}
\begin{center}
\resizebox{\columnwidth}{!}{%
\begin{tabular}{lrrr}
\multicolumn{1}{c}{\textbf{Model}} & \multicolumn{1}{c}{\textbf{Jaccard}} & \multicolumn{1}{c}{\textbf{Spearman}} & \multicolumn{1}{c}{\textbf{Pearson}}
\\ \hline \\
Integrated Gradients & $0.855 \pm 0.028$ & $0.758 \pm 0.038$ & $0.969 \pm 0.009$ \\
Noise Tunnel         & $0.954 \pm 0.031$ & $0.752 \pm 0.038$ & $0.974 \pm 0.007$ \\
Deep Lift            & $0.976 \pm 0.034$ & $0.760 \pm 0.037$ & $0.971 \pm 0.008$ \\
Gradient SHAP        & $0.939 \pm 0.041$ & $0.750 \pm 0.041$ & $0.975 \pm 0.006$ \\
Feature Ablation     & $0.956 \pm 0.050$ & $0.759 \pm 0.037$ & $0.980 \pm 0.005$ \\
FSL                  & $\mathbf{1.000 \pm 0.000}$ & $\mathbf{0.805 \pm 0.030}$ & $\mathbf{0.980 \pm 0.003}$ \\
Post-hoc FSL         & $0.759 \pm 0.052$ & $0.759 \pm 0.030$ & $0.770 \pm 0.027$ \\
\end{tabular}
}
\end{center}
\end{table}

\begin{figure}[ht]
    \centering
    \begin{subfigure}[b]{0.2\textwidth}
         \centering
         \includegraphics[width=\textwidth]{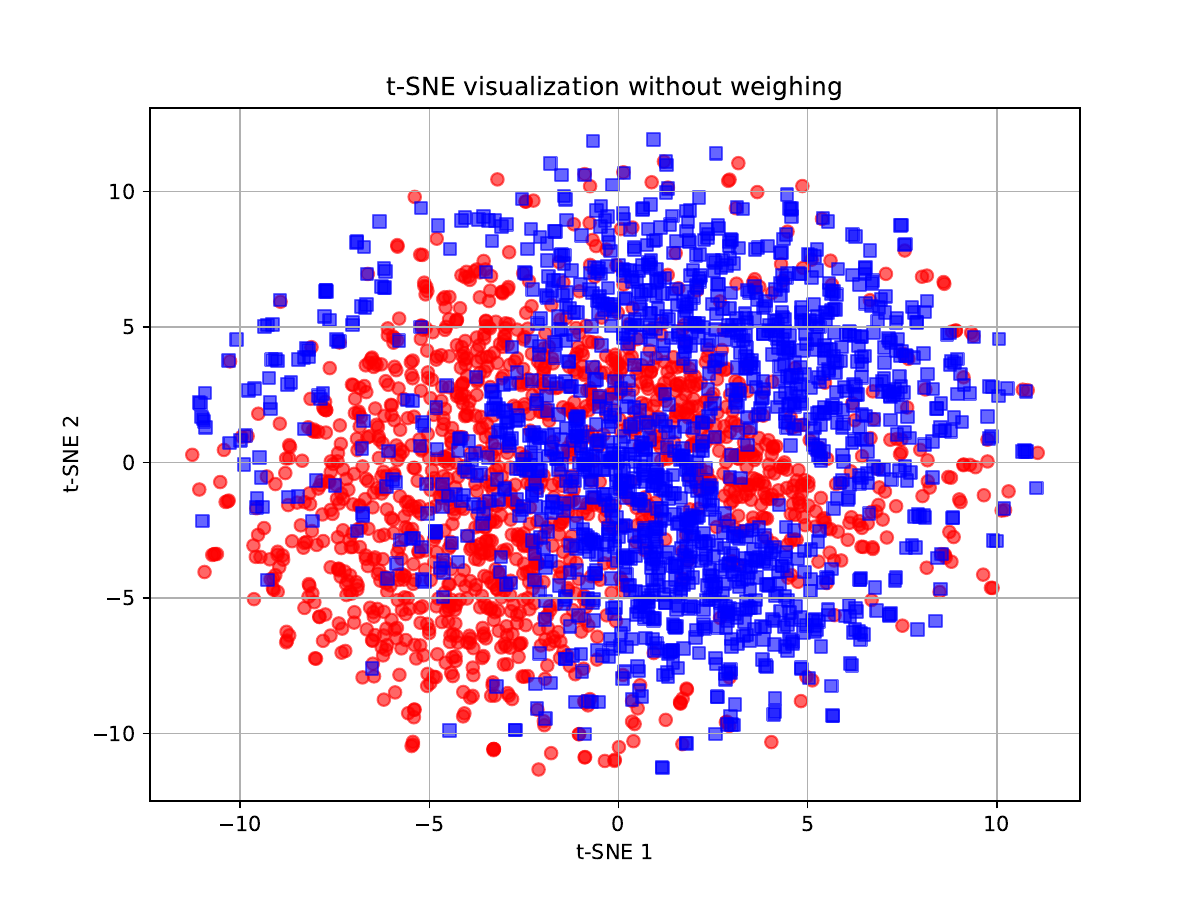}
         \caption{Without weights}
         \label{subfig:wtsne_tabfn_without_weights}
     \end{subfigure}
     \begin{subfigure}[b]{0.2\textwidth}
         \centering
         \includegraphics[width=\textwidth]{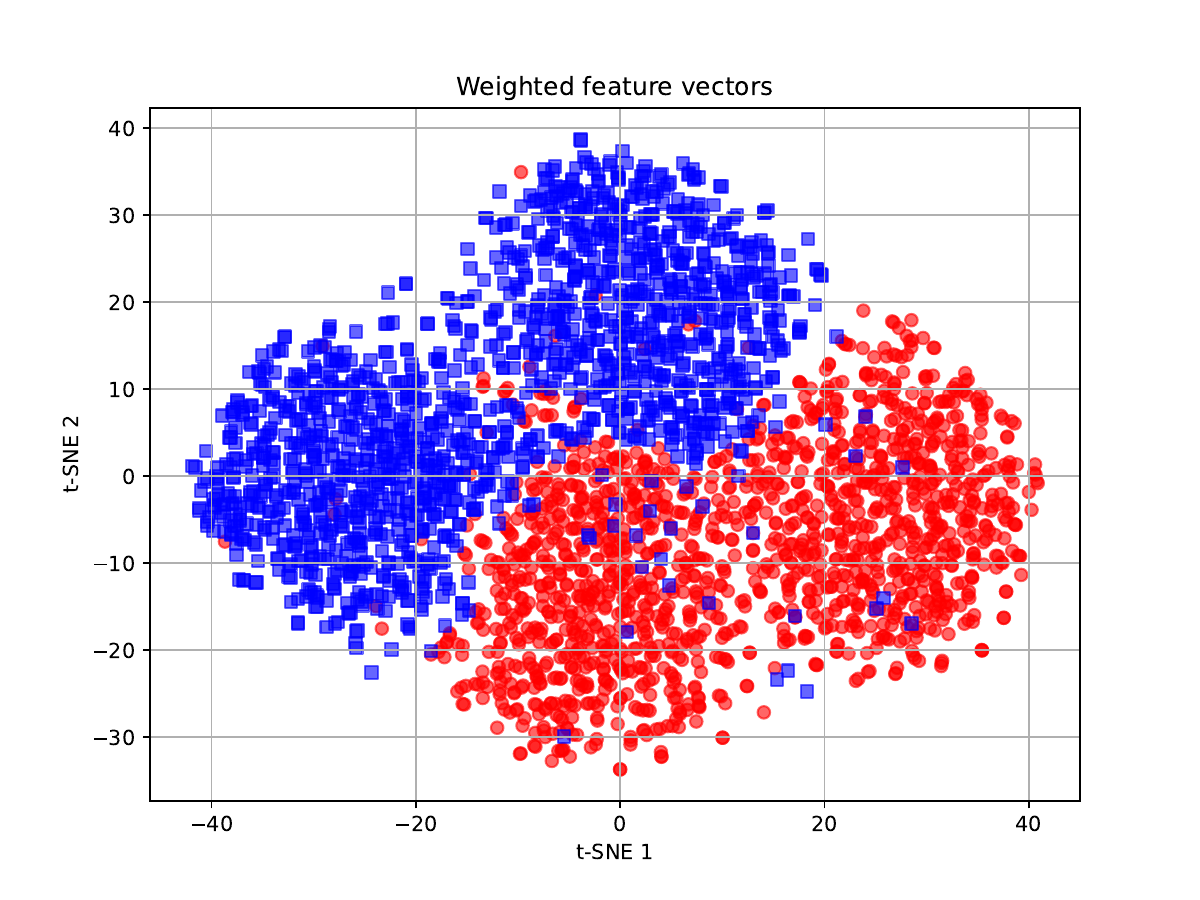}
         \caption{FSL post-hoc}
         \label{subfig:wtsne_tabfn_fsl_post_hoc}
     \end{subfigure}
     \begin{subfigure}[b]{0.2\textwidth}
         \centering
         \includegraphics[width=\textwidth]{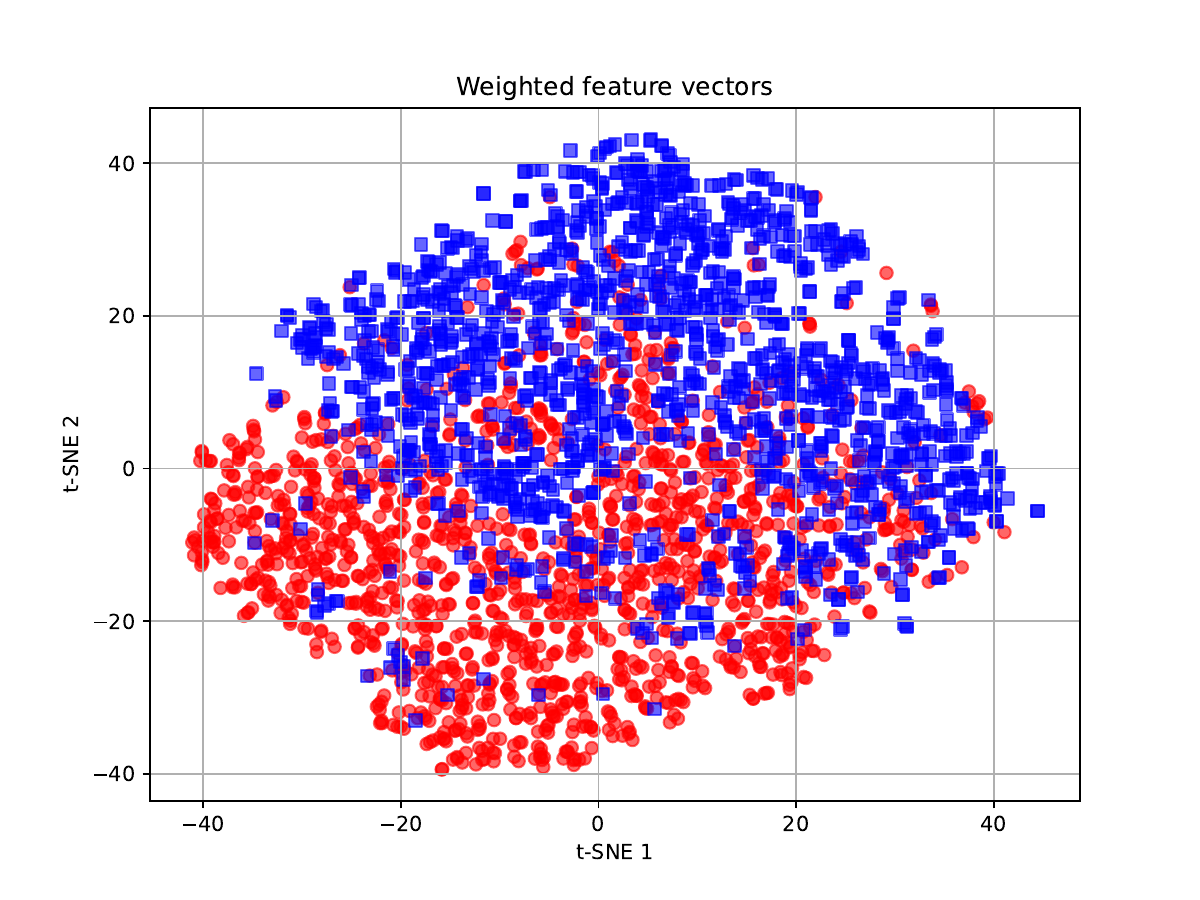}
         \caption{Kernel SHAP}
         \label{subfig:wtsne_tabfn_shap}
     \end{subfigure}
    \caption[Weighted t-SNE]
    {\textbf{Weighted t-SNE} of the post-hoc FSL and Kernel SHAP methods using the TabPFN and the SynthA dataset.
    }
    \label{fig:wtsne_tabfn}
\end{figure}

To complement our experiments, we used TabPFN \cite{hollmann2022tabpfn}, a pre-trained Transformer-based model for supervised classification on small tabular datasets that operates without hyperparameter tuning and achieves competitive performance against state-of-the-art classification methods.
Due to source code compatibility, comparisons were made using the original Kernel SHAP implementation. 
Both post-hoc techniques successfully identified the most relevant features, yielding identical PSFI and PIFS scores of $0.966$ when evaluating the top-$30$ ranked features, which ideally should include all, and only, the informative features.
However, post-hoc FSL visualization outperformed both weighted t-SNE using Kernel SHAP weights and standard t-SNE (\autoref{fig:wtsne_tabfn}), achieving a higher silhouette score of $0.316$ compared to $0.186$ and $0.070$, respectively.

\subsection{Results for Real-world Datasets}
\label{subsec:realresults}
\begin{figure}[ht]
    \centering
    \begin{subfigure}[b]{0.2\textwidth}
         \centering
         \includegraphics[width=\textwidth]{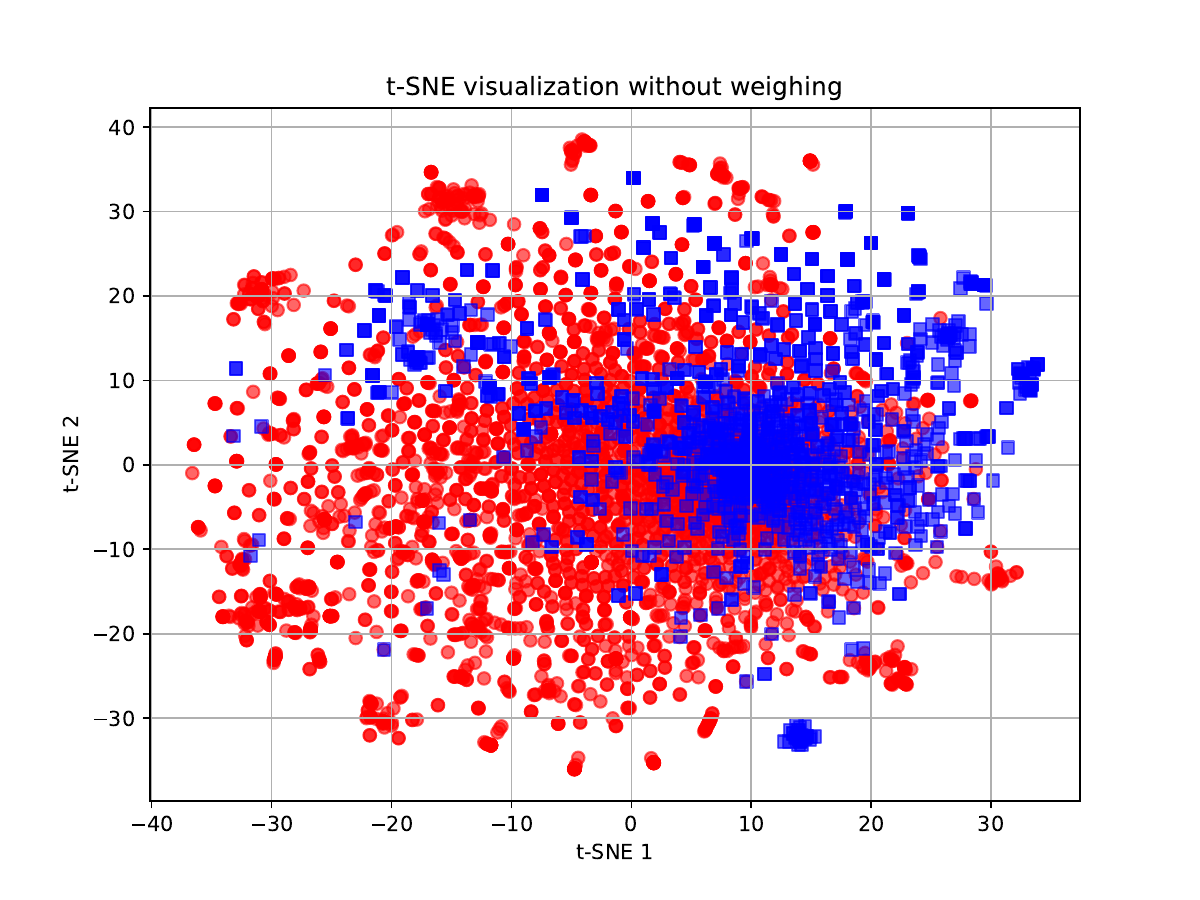}
         \caption{Without weights}
         \label{subfig:wtsne_Spam_without_weights}
     \end{subfigure}
     \begin{subfigure}[b]{0.2\textwidth}
         \centering
         \includegraphics[width=\textwidth]{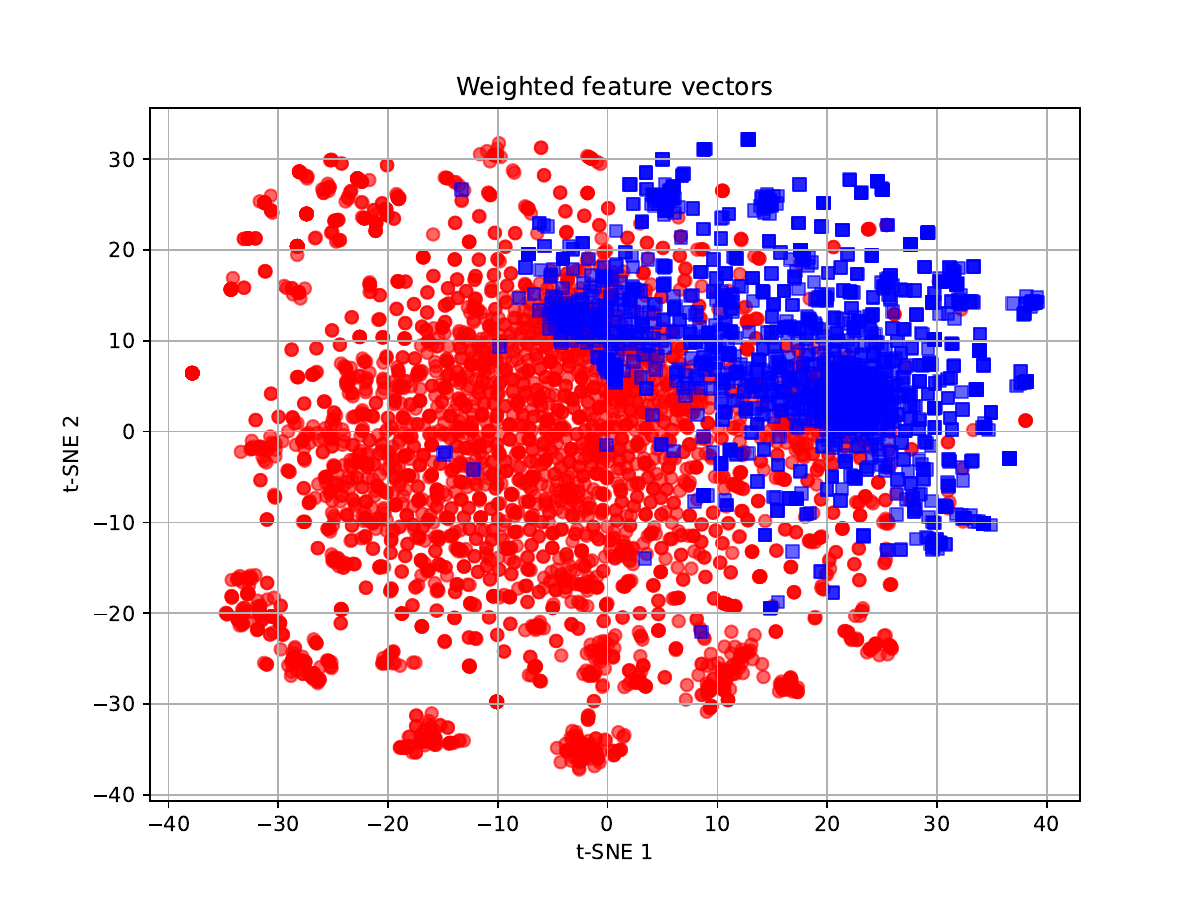}
         \caption{Integrated Gradients}
         \label{subfig:wtsne_Spam_integratedgradients_weights}
     \end{subfigure}
     \begin{subfigure}[b]{0.2\textwidth}
         \centering
         \includegraphics[width=\textwidth]{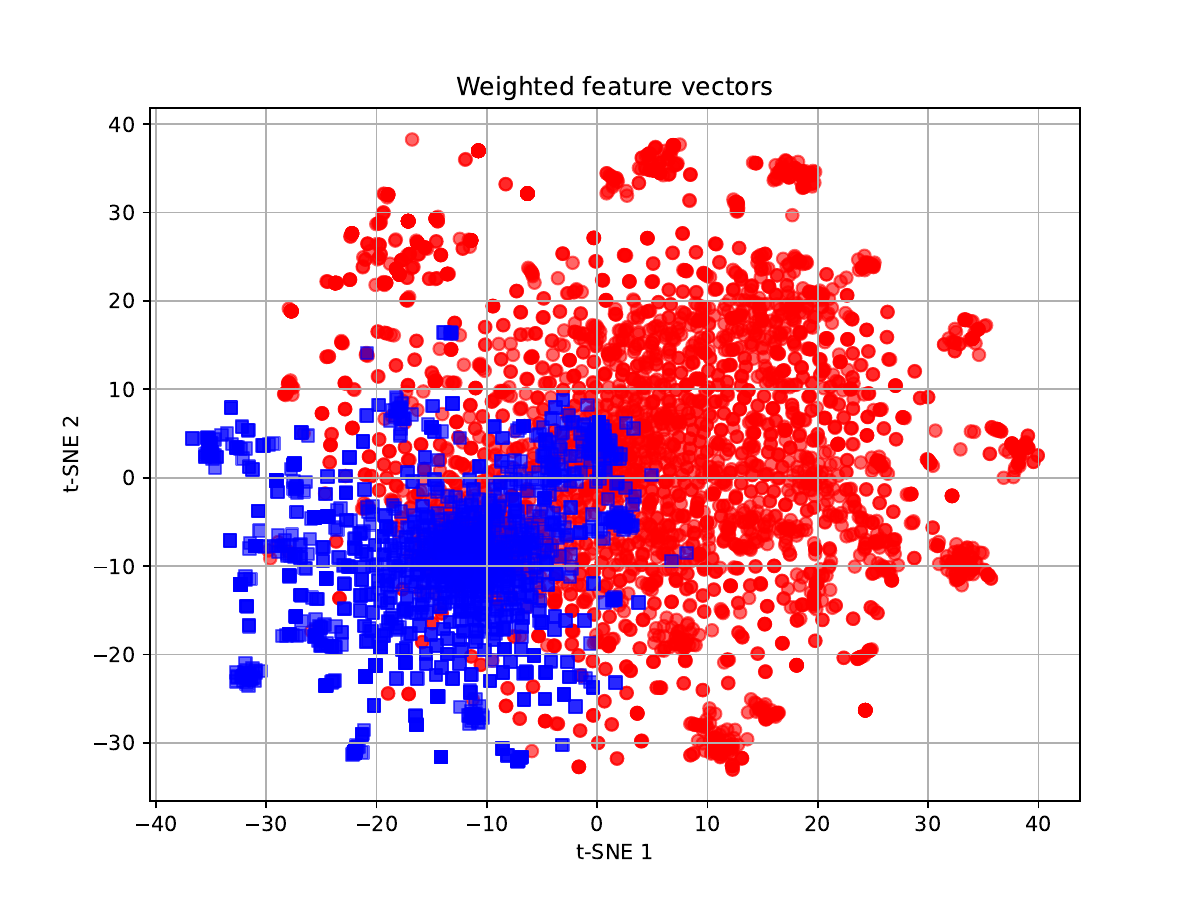}
         \caption{Noise Tunnel}
         \label{subfig:wtsne_Spam_noisetunnel_weights}
     \end{subfigure}
     \begin{subfigure}[b]{0.2\textwidth}
         \centering
         \includegraphics[width=\textwidth]{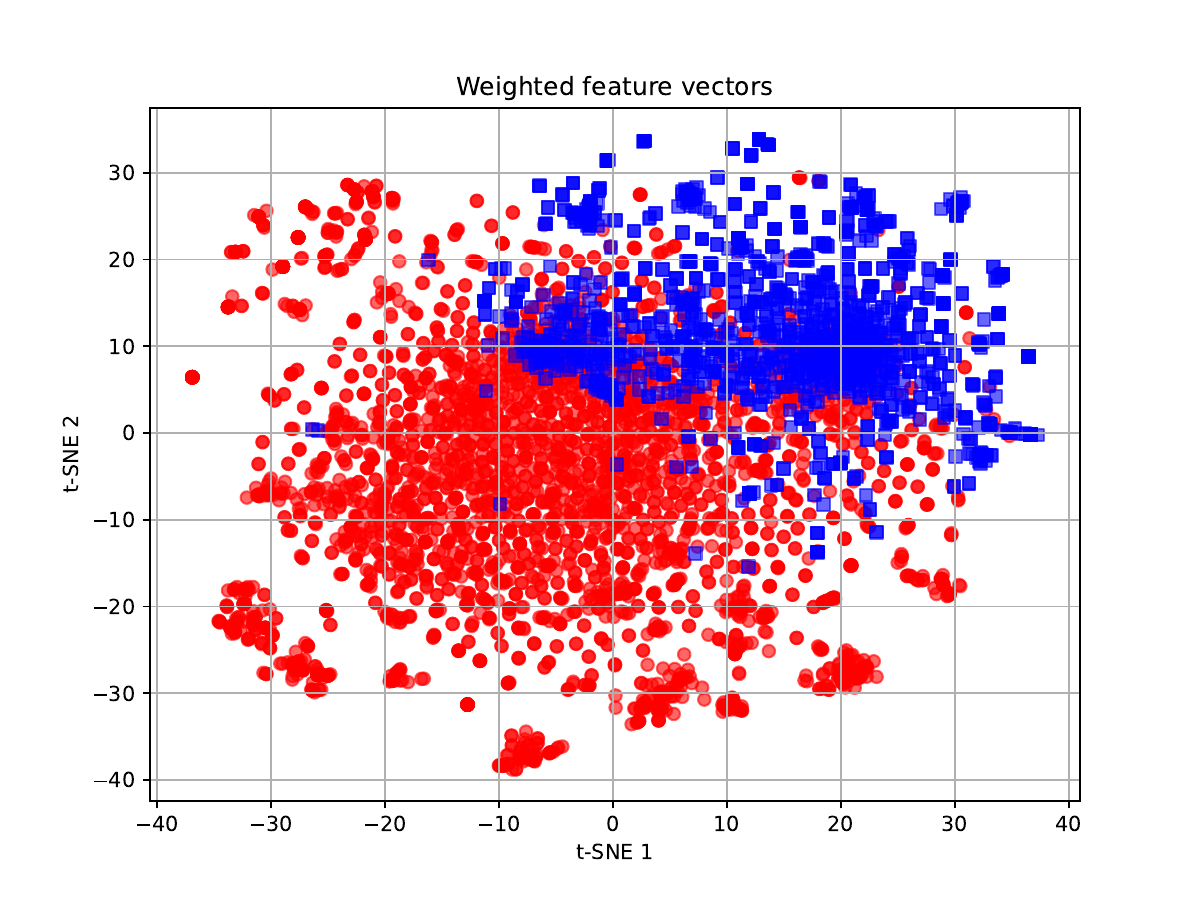}
         \caption{Deep Lift}
         \label{subfig:wtsne_Spam_deeplift_weights}
     \end{subfigure}
     \begin{subfigure}[b]{0.2\textwidth}
         \centering
         \includegraphics[width=\textwidth]{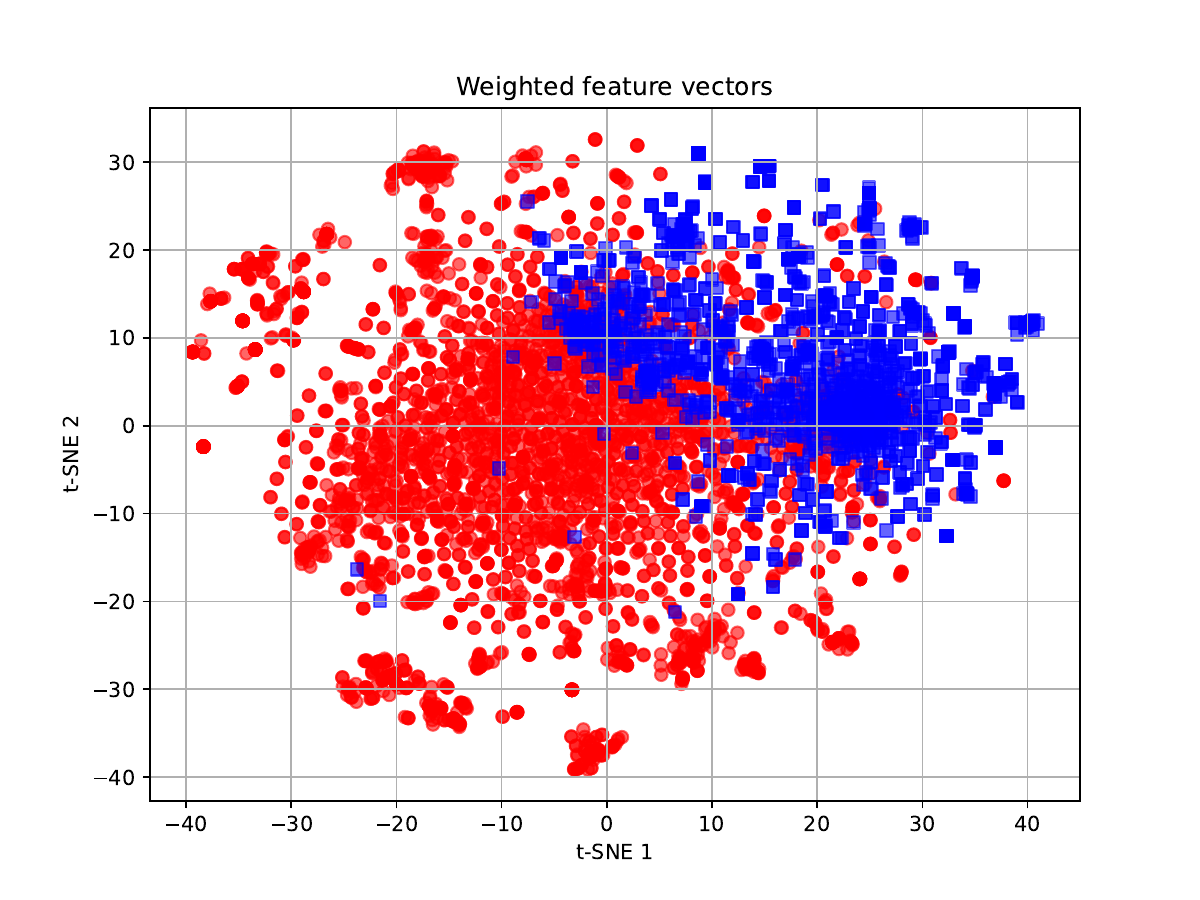}
         \caption{Gradient SHAP}
         \label{subfig:wtsne_Spam_gradientshap_weights}
     \end{subfigure}
     \begin{subfigure}[b]{0.2\textwidth}
         \centering
         \includegraphics[width=\textwidth]{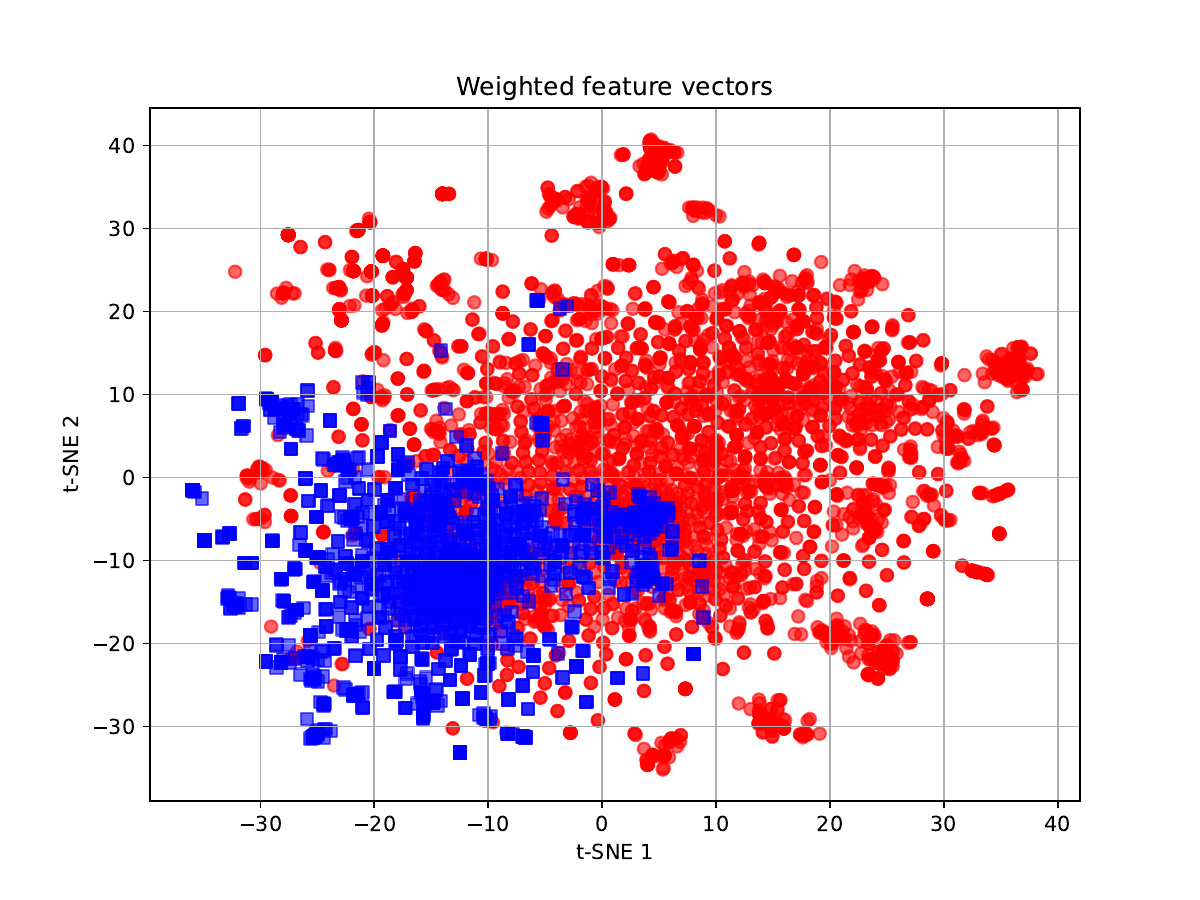}
         \caption{Feature Ablation}
         \label{subfig:wtsne_Spam_featureablation_weights}
     \end{subfigure}
     \begin{subfigure}[b]{0.2\textwidth}
         \centering
         \includegraphics[width=\textwidth]{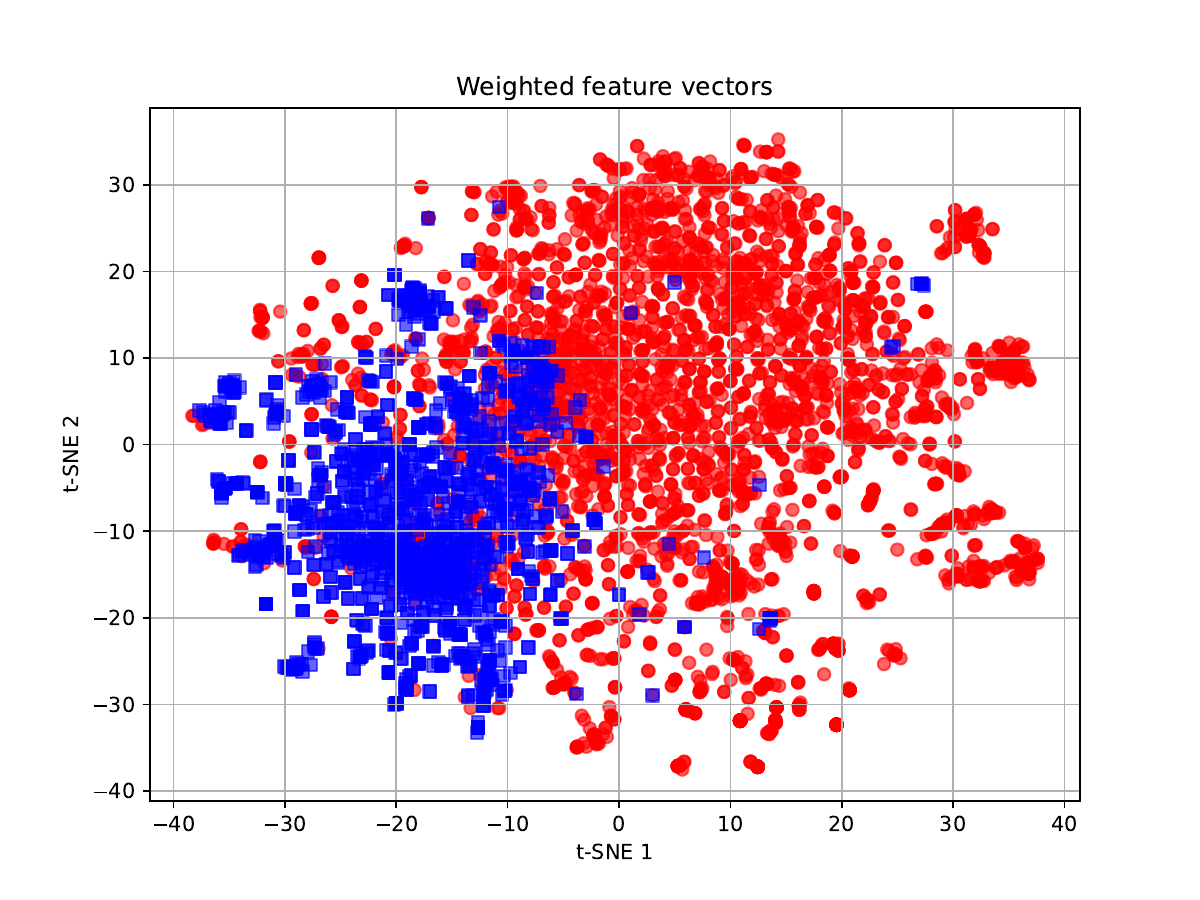}
         \caption{FSL}
         \label{subfig:wtsne_Spam_fsl_weights}
     \end{subfigure}
     \begin{subfigure}[b]{0.2\textwidth}
         \centering
         \includegraphics[width=\textwidth]{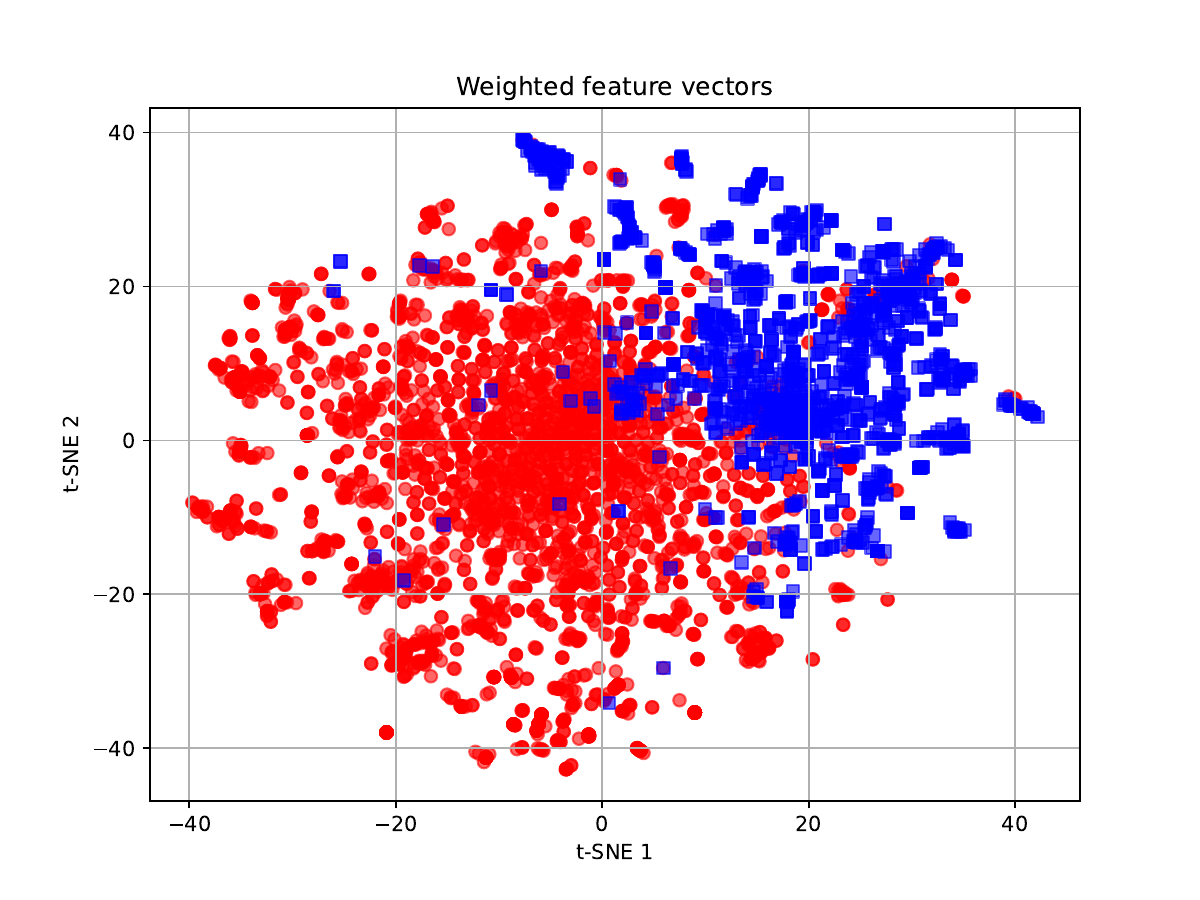}
         \caption{Post-hoc FSL}
         \label{subfig:wtsne_Spam_posthocfsl_weights}
     \end{subfigure}
    \caption[Weighted t-SNE]
    {\textbf{Weighted t-SNE} 
         of all feature weighing metrics using the spam dataset.
    }
    \label{fig:wtsne_spam}
\end{figure}

We evaluated the post-hoc FSL on real-world datasets (Section \ref{subsubsec:realdatasets}), including spam email detection and high-dimensional, low-sample-size microarray analysis.
The results for the spam dataset are presented in \autoref{table:performance_real_world_datasets}, \ref{table:real_world_silhouette} and \autoref{fig:wtsne_spam}. 
As shown in \autoref{table:performance_real_world_datasets}, the baseline and post-hoc FSL models achieved the highest performance metrics.
The baseline model recorded the best average scores across accuracy, F1 score, recall, and precision. 
While the baseline model demonstrated the top performance, post-hoc FSL obtained comparable results and outperformed FSL. 
In terms of visual separability metrics, the FSL post-hoc yielded the best results, achieving a higher average silhouette coefficient than both the FSL and the other post-hoc methods (\autoref{table:real_world_silhouette} and \autoref{fig:wtsne_spam}).
However, in terms of stability, both FSL variants underperformed compared to the other post-hoc approaches.
The Jaccard index, computed using the top-$100$ weighted features, ranged from $0.382$ to $0.486$ across all methods, indicating relatively consistent feature selection across runs.
In contrast, Spearman rank correlation was substantially lower for FSL and post-hoc FSL ($0.157$ and $0.312$), while the remaining post-hoc methods achieved averages between $0.553$ and $0.741$.
Pearson correlation was generally higher across all approaches ($0.614$–$0.818$), likely reflecting the large number of consistently irrelevant features with near-zero weights.
Although our proposed FSL achieved comparable Jaccard and Pearson scores, it exhibited lower overall stability than other post-hoc methods. Nevertheless, it consistently assigned meaningful weights to a subset of highly relevant features, suggesting that while the global ranking may vary, the selected features remain informative.

\begin{table}[ht]
    \centering
    \caption{
    \textbf{Prediction performance for real-world datasets.}
    Bold indicates the highest average; best statistically significant results ($p < 0.01$) from Kruskal-Wallis and Dunn’s tests are marked in orange.
    }
    \label{table:performance_real_world_datasets}
    
    \begin{subtable}[t]{\columnwidth}
        \centering
        \caption{Spam dataset}
        \resizebox{\columnwidth}{!}{%
        \begin{tabular}{lrrrr}
        \multicolumn{1}{c}{\textbf{Model}} & \multicolumn{1}{c}{\textbf{F1 Score}} & \multicolumn{1}{c}{\textbf{Accuracy}} & \multicolumn{1}{c}{\textbf{Precision}} & \multicolumn{1}{c}{\textbf{Recall}}
        \\ \hline \\
        Baseline & \cellcolor{orange!20} $\mathbf{0.976 \pm 0.005}$ & \cellcolor{orange!20} $\mathbf{0.976 \pm 0.005}$ & \cellcolor{orange!20} $\mathbf{0.977 \pm 0.005}$ & \cellcolor{orange!20} $\mathbf{0.976 \pm 0.005}$ \\
        FSL & $0.968 \pm 0.004$ & $0.968 \pm 0.004$ & $0.969 \pm 0.004$ & $0.968 \pm 0.004$ \\
        FSL post-hoc & \cellcolor{orange!20} $0.973 \pm 0.003$ & \cellcolor{orange!20} $0.973 \pm 0.003$ & \cellcolor{orange!20} $0.974 \pm 0.003$ & \cellcolor{orange!20} $0.973 \pm 0.003$ \\
        \end{tabular}
        }
    \end{subtable}

    \begin{subtable}[t]{\columnwidth}
        \centering
        \caption{Liver dataset}
        \resizebox{\columnwidth}{!}{%
        \begin{tabular}{lrrrr}
        \multicolumn{1}{c}{\bf Model} &\multicolumn{1}{c}{\bf F1 Score} &\multicolumn{1}{c}{\bf Accuracy} &\multicolumn{1}{c}{\bf Precision} &\multicolumn{1}{c}{\bf Recall}
        \\ \hline \\
        Baseline & \cellcolor{orange!20} $0.800 \pm 0.076$ & \cellcolor{orange!20} $0.800 \pm 0.076$ & \cellcolor{orange!20} $0.858 \pm 0.079$ & \cellcolor{orange!20} $0.800 \pm 0.076$ \\
        FSL & \cellcolor{orange!20} $\mathbf{0.840 \pm 0.083}$ & \cellcolor{orange!20} $\mathbf{0.840 \pm 0.083}$ & \cellcolor{orange!20} $\mathbf{0.893 \pm 0.055}$ & \cellcolor{orange!20} $\mathbf{0.840 \pm 0.083}$ \\
        Post-hoc FSL & \cellcolor{orange!20} $0.827 \pm 0.070$ & \cellcolor{orange!20} $0.827 \pm 0.070$ & \cellcolor{orange!20} $0.884 \pm 0.047$ & \cellcolor{orange!20} $0.827 \pm 0.070$ \\
        \end{tabular}
        }
    \end{subtable}
     
    \begin{subtable}[t]{\columnwidth}
        \centering
        \caption{Breast dataset}
        \resizebox{\columnwidth}{!}{%
        \begin{tabular}{lrrrr}
        \multicolumn{1}{c}{\bf Model} &\multicolumn{1}{c}{\bf F1 Score} &\multicolumn{1}{c}{\bf Accuracy} &\multicolumn{1}{c}{\bf Precision} &\multicolumn{1}{c}{\bf Recall}
        \\ \hline \\
        Baseline & \cellcolor{orange!20} $0.942 \pm 0.024$ & \cellcolor{orange!20} $0.945 \pm 0.021$ & \cellcolor{orange!20} $0.955 \pm 0.024$ & \cellcolor{orange!20} $0.945 \pm 0.021$ \\
        FSL & \cellcolor{orange!20} $0.955 \pm 0.033$ & \cellcolor{orange!20} $0.958 \pm 0.030$ & \cellcolor{orange!20} $0.964 \pm 0.030$ & \cellcolor{orange!20} $0.958 \pm 0.030$ \\
        FSL post-hoc & \cellcolor{orange!20} $\mathbf{0.970 \pm 0.033}$ & \cellcolor{orange!20} $\mathbf{0.970 \pm 0.032}$ & \cellcolor{orange!20} $\mathbf{0.978 \pm 0.024}$ & \cellcolor{orange!20} $\mathbf{0.970 \pm 0.032}$ \\
        \end{tabular}
        }
    \end{subtable}

    \label{tab:stacked_results}
\end{table}

\begin{table}[ht]
\caption{
\textbf{Silhouette score for real-world dataset.}
Bold indicates the highest average; best statistically significant results ($p < 0.01$) from Kruskal-Wallis and Dunn’s tests are marked in orange.
}
\label{table:real_world_silhouette}
\begin{center}
\resizebox{\columnwidth}{!}{%
\begin{tabular}{lrrr}
\multicolumn{1}{c}{\textbf{Model}} & \multicolumn{1}{c}{\textbf{Spam}} & \multicolumn{1}{c}{\textbf{Liver}} & \multicolumn{1}{c}{\textbf{Breast}}
\\ \hline \\
None                 & $0.090 \pm 0.000$ & $0.066 \pm 0.000$ & $0.192 \pm 0.000$ \\
Integrated Gradients & \cellcolor{orange!20} $0.202 \pm 0.031$ & $0.170 \pm 0.031$ & $0.451 \pm 0.036$ \\
Noise Tunnel         & \cellcolor{orange!20} $0.189 \pm 0.034$ & $0.178 \pm 0.044$ & $0.447 \pm 0.045$ \\
Deep Lift            & \cellcolor{orange!20} $0.202 \pm 0.025$ & $0.172 \pm 0.049$ & $0.449 \pm 0.030$ \\
Gradient SHAP        & \cellcolor{orange!20} $0.209 \pm 0.022$ & $0.199 \pm 0.042$ & $0.450 \pm 0.040$ \\
Feature Ablation     & \cellcolor{orange!20} $0.196 \pm 0.029$ & $0.181 \pm 0.030$ & $0.450 \pm 0.033$ \\
FSL                  & \cellcolor{orange!20} $0.234 \pm 0.013$ & \cellcolor{orange!20} $0.574 \pm 0.141$ & \cellcolor{orange!20} $\mathbf{0.647 \pm 0.021}$ \\
Post-hoc FSL         & \cellcolor{orange!20} $\mathbf{0.235 \pm 0.069}$ & \cellcolor{orange!20} $\mathbf{0.666 \pm 0.012}$ & \cellcolor{orange!20} $0.511 \pm 0.078$ \\
\end{tabular}
}
\end{center}
\end{table}

\begin{table}[ht]
\caption{
\textbf{Results for stability metrics using the Liver dataset.} 
In bold is the highest average of each metric. Jaccard was calculated using the top-$100$ weighted features.}
\label{table:liverstabilityresults}
\begin{center}
\resizebox{\columnwidth}{!}{%
\begin{tabular}{lrrr}
\multicolumn{1}{c}{\textbf{Model}} & \multicolumn{1}{c}{\textbf{Jaccard}} & \multicolumn{1}{c}{\textbf{Spearman}} & \multicolumn{1}{c}{\textbf{Pearson}}
\\ \hline \\
Integrated Gradients & $0.263 \pm 0.078$ & $\mathbf{0.649 \pm 0.091}$ & $0.739 \pm 0.081$ \\
Noise Tunnel         & $0.226 \pm 0.102$ & $0.647 \pm 0.092$ & $0.728 \pm 0.088$ \\
Deep Lift            & $0.263 \pm 0.077$ & $\mathbf{0.649 \pm 0.091}$ & $0.739 \pm 0.081$ \\
Gradient SHAP        & $0.122 \pm 0.047$ & $0.599 \pm 0.098$ & $0.666 \pm 0.084$ \\
Feature Ablation     & $0.261 \pm 0.077$ & $\mathbf{0.649 \pm 0.092}$ & $0.739 \pm 0.081$ \\
FSL                  & $0.174 \pm 0.064$ & $0.017 \pm 0.016$ & $0.454 \pm 0.180$ \\
Post-hoc FSL         & $\mathbf{0.472 \pm 0.086}$ & $0.090 \pm 0.014$ & $\mathbf{0.832 \pm 0.058}$ \\
\end{tabular}
}
\end{center}
\end{table}

On the Liver and Breast datasets, post-hoc FSL achieved predictive performance comparable to the baseline and original FSL (\autoref{table:performance_real_world_datasets}), while consistently producing more meaningful feature attributions. 
Both FSL variants outperformed state-of-the-art post-hoc methods in silhouette scores (\autoref{table:real_world_silhouette}), with post-hoc FSL peaking on the Liver dataset ($0.666$) and original FSL on the Breast dataset ($0.647$), demonstrating their ability to create well-defined class clusters. 
Regarding stability (\autoref{table:liverstabilityresults}), post-hoc FSL yielded the highest Jaccard and Pearson scores on the Liver dataset, although both FSL variants showed lower Spearman scores. 
Conversely, both methods exhibited higher instability across all metrics on the Breast dataset, likely because the FSL architecture struggles to capture local, class-specific relationships across its six classes.

\subsection{Feature Erasure}

We further evaluated feature relevance using Feature Erasure, which measures performance impact by zeroing out the top-$n$ features ranked by post-hoc weights and later performing inference on the baseline model. 
We progressively erased the top-$1$ to $50$ features for the XOR and SynthA datasets, and top-$100$ to $4000$ features for the Spam dataset. 
After masking the top-$2$ for XOR, top-$30$ for SynthA and top-$2000$ for Spam the baseline model's F1-Score and accuracy significantly degraded. 
All methods, including post-hoc FSL, exhibited similar performance degradation, demonstrating their effectiveness in selecting relevant features.

\section{Conclusion}

This paper explores the development of a variation of the Feature Selection Layer (FSL) as a feature weighing method to improve the interpretability of pre-trained Deep Neural Networks. 
Our results demonstrate that the proposed method effectively isolates relevant features in high-dimensional datasets, matching or outperforming state-of-the-art post-hoc feature weighting methods.
Unlike traditional methods, Post-hoc FSL can be integrated directly into the model architecture for subsequent inferences. 
By driving irrelevant feature weights toward zero, the method systematically excludes non-contributory inputs, thereby
increasing reliability.
Experimental results further indicate that coupling Post-hoc FSL typically yields a positive impact on the model's predictive performance.

\subsection{Limitations and future work}


While effective for binary classification, our proposed method underperforms in multi-class settings, as it fails to capture local and class-specific feature relationships. Furthermore, our proposed method is specifically designed for tabular data, in which each feature has a fixed meaning within the dataset; however, this assumption does not hold for images, where the meaning of each pixel can vary across samples. Future work should focus on adapting the method to overcome these limitations and extend its interpretability to such domains.

 
\subsection*{Reproducibility Statement}

\textbf{Paper source code} is available at \href{https://github.com/joaosuwa/FSL_post-hoc}{https://github.com/joaosuwa/FSL\_post-hoc} with all necessary dependencies, seeds and steps to reproduce each of the experiments presented in this paper.
The original code of the weighted t-SNE \cite{grisci2025assessing} can be found at   \href{https://github.com/sbcblab/weighted_tSNE}{https://github.com/sbcblab/weighted\_tSNE}.
\textbf{CuMiDa} datasets \cite{cumida:2019} can be found at \href{https://sbcb.inf.ufrgs.br/cumida}{https://sbcb.inf.ufrgs.br/cumida} and \textbf{Spam} dataset \cite{balaka2020spam} can be found at \href{https://www.kaggle.com/datasets/balaka18/email-spam-classification-dataset-csv}{https://www.kaggle.com/datasets/balaka18/email-spam-classification-dataset-csv}.
Both \textbf{XOR} and \textbf{SynthA} dataset are available at the same repository of the paper source code.
\textbf{TabPFN source code} is available at \href{https://github.com/PriorLabs/TabPFN}{https://github.com/PriorLabs/TabPFN} and \textbf{Kernel SHAP documentation} is available at \href{https://shap.readthedocs.io/en/latest/}{https://shap.readthedocs.io/en/latest/}.




\bibliographystyle{IEEEtran}
\bibliography{references}

@article{badawy2023healthcare,
  title={Healthcare predictive analytics using machine learning and deep learning techniques: a survey},
  author={Badawy, Mohammed and Ramadan, Nagy and Hefny, Hesham Ahmed},
  journal={Journal of Electrical Systems and Information Technology},
  volume={10},
  number={1},
  pages={40},
  year={2023},
  publisher={Springer}
}

@article{KHAN2023100026,
title = {Exploring the frontiers of deep learning and natural language processing: A comprehensive overview of key challenges and emerging trends},
journal = {Natural Language Processing Journal},
volume = {4},
pages = {100026},
year = {2023},
issn = {2949-7191},
doi = {https://doi.org/10.1016/j.nlp.2023.100026},
url = {https://www.sciencedirect.com/science/article/pii/S2949719123000237},
author = {Wahab Khan and Ali Daud and Khairullah Khan and Shakoor Muhammad and Rafiul Haq},
}

@article{grisci2024use,
  title={The use of gene expression datasets in feature selection research: 20 years of inherent bias?},
  author={Grisci, Bruno I and Feltes, Bruno C{\'e}sar and de Faria Poloni, Joice and Narloch, Pedro H and Dorn, Marcio},
  journal={Wiley Interdisciplinary Reviews: Data Mining and Knowledge Discovery},
  volume={14},
  number={2},
  pages={e1523},
  year={2024},
  publisher={Wiley Online Library}
}

@ARTICLE{Unraveling_the_Black_Box:2024,
  author={Murad, Nafeesa Yousuf and Hasan, Mohd Hilmi and Azam, Muhammad Hamza and Yousuf, Nadia and Yalli, Jameel Shehu},
  journal={IEEE Access}, 
  title={Unraveling the Black Box: A Review of Explainable Deep Learning Healthcare Techniques}, 
  year={2024},
  volume={12},
  number={},
  pages={66556-66568},
  doi={10.1109/ACCESS.2024.3398203}}

@inproceedings{gui2019afs,
  title={AFS: An attention-based mechanism for supervised feature selection},
  author={Gui, Ning and Ge, Danni and Hu, Ziyin},
  booktitle={Proceedings of the AAAI conference on artificial intelligence},
  volume={33},
  number={01},
  pages={3705--3713},
  year={2019}
}

@misc{balaka2020spam,
  author = {Balaka},
  title = {Email Spam Classification Dataset (CSV)},
  year = {2020},
  howpublished = {\url{https://www.kaggle.com/datasets/balaka18/email-spam-classification-dataset-csv}},
  note = {Accessed: 2025-08-20}
}

@misc{kokhlikyan2020captumunifiedgenericmodel,
      title={Captum: A unified and generic model interpretability library for PyTorch}, 
      author={Narine Kokhlikyan and Vivek Miglani and Miguel Martin and Edward Wang and Bilal Alsallakh and Jonathan Reynolds and Alexander Melnikov and Natalia Kliushkina and Carlos Araya and Siqi Yan and Orion Reblitz-Richardson},
      year={2020},
      eprint={2009.07896},
      archivePrefix={arXiv},
      primaryClass={cs.LG},
      url={https://arxiv.org/abs/2009.07896}, 
}

@article{BARBIERI2024123667,
title = {Analysis and comparison of feature selection methods towards performance and stability},
journal = {Expert Systems with Applications},
volume = {249},
pages = {123667},
year = {2024},
issn = {0957-4174},
doi = {https://doi.org/10.1016/j.eswa.2024.123667},
url = {https://www.sciencedirect.com/science/article/pii/S0957417424005335},
author = {Matheus Cezimbra Barbieri and Bruno Iochins Grisci and Márcio Dorn}
}

@book{prince2023understanding,
        author = "Simon J.D. Prince",
        title = "Understanding Deep Learning",
        publisher = "The MIT Press",
        year = 2023,
        url = "http://udlbook.com"
    }

@inproceedings{molnar2020interpretable,
  title={Interpretable machine learning--a brief history, state-of-the-art and challenges},
  author={Molnar, Christoph and Casalicchio, Giuseppe and Bischl, Bernd},
  booktitle={Joint European conference on machine learning and knowledge discovery in databases},
  pages={417--431},
  year={2020},
  organization={Springer}
}

@article{figueroa2021towards,
  title={Towards interpretable deep learning: a feature selection framework for prognostics and health management using deep neural networks},
  author={Figueroa Barraza, Joaqu{\'\i}n and L{\'o}pez Droguett, Enrique and Martins, Marcelo Ramos},
  journal={Sensors},
  volume={21},
  number={17},
  pages={5888},
  year={2021},
  publisher={MDPI}
}

@article{tahir2007simultaneous,
  title={Simultaneous feature selection and feature weighting using Hybrid Tabu Search/K-nearest neighbor classifier},
  author={Tahir, Muhammad Atif and Bouridane, Ahmed and Kurugollu, Fatih},
  journal={Pattern Recognition Letters},
  volume={28},
  number={4},
  pages={438--446},
  year={2007},
  publisher={Elsevier}
}

@article{miao2016survey,
  title={A survey on feature selection},
  author={Miao, Jianyu and Niu, Lingfeng},
  journal={Procedia computer science},
  volume={91},
  pages={919--926},
  year={2016},
  publisher={Elsevier}
}

@inproceedings{sundararajan2017axiomatic,
  title={Axiomatic attribution for deep networks},
  author={Sundararajan, Mukund and Taly, Ankur and Yan, Qiqi},
  booktitle={International conference on machine learning},
  pages={3319--3328},
  year={2017},
  organization={PMLR}
}

@inproceedings{shrikumar2017learning,
  title={Learning important features through propagating activation differences},
  author={Shrikumar, Avanti and Greenside, Peyton and Kundaje, Anshul},
  booktitle={International conference on machine learning},
  pages={3145--3153},
  year={2017},
  organization={PMlR}
}

@article{lundberg2017unified,
  title={A unified approach to interpreting model predictions},
  author={Lundberg, Scott M and Lee, Su-In},
  journal={Advances in neural information processing systems},
  volume={30},
  year={2017}
}

@article{cumida:2019,
	title        = {CuMiDa: An Extensively Curated Microarray Database for Benchmarking and Testing of Machine Learning Approaches in Cancer Research},
	author       = {Feltes, B.C. and Chandelier, E. B. and Grisci, B. I. and Dorn, M.},
	year         = 2019,
	journal      = {Journal of Computational Biology},
	volume       = 26,
	number       = 4,
	pages        = {376--386},
	doi          = {10.1089/cmb.2018.0238}
}

@article{maaten2008visualizing,
  title={Visualizing data using t-SNE},
  author={Maaten, Laurens van der and Hinton, Geoffrey},
  journal={Journal of machine learning research},
  volume={9},
  number={Nov},
  pages={2579--2605},
  year={2008}
}

@article{grisci2025assessing,
  title={Assessing feature scorer results on high-dimensional datasets with t-SNE},
  author={Grisci, Bruno Iochins and Inostroza-Ponta, Mario and Dorn, M{\'a}rcio},
  journal={Neurocomputing},
  pages={130561},
  year={2025},
  publisher={Elsevier}
}

@inproceedings{awada2012review,
  title={A review of the stability of feature selection techniques for bioinformatics data},
  author={Awada, Wael and Khoshgoftaar, Taghi M and Dittman, David and Wald, Randall and Napolitano, Amri},
  booktitle={2012 IEEE 13th International conference on information reuse \& integration (IRI)},
  pages={356--363},
  year={2012},
  organization={IEEE}
}

@article{ang2015supervised,
  title={Supervised, unsupervised, and semi-supervised feature selection: a review on gene selection},
  author={Ang, Jun Chin and Mirzal, Andri and Haron, Habibollah and Hamed, Haza Nuzly Abdull},
  journal={IEEE/ACM transactions on computational biology and bioinformatics},
  volume={13},
  number={5},
  pages={971--989},
  year={2015},
  publisher={IEEE}
}

@article{hollmann2022tabpfn,
  title={Tabpfn: A transformer that solves small tabular classification problems in a second},
  author={Hollmann, Noah and M{\"u}ller, Samuel and Eggensperger, Katharina and Hutter, Frank},
  journal={arXiv preprint arXiv:2207.01848},
  year={2022}
}

\end{document}